\documentclass[11pt,a4paper]{article}
\usepackage[hyperref]{emnlp2020}
\usepackage{times}
\usepackage{latexsym}

\usepackage[ruled,vlined]{algorithm2e}
\usepackage{multirow}
\usepackage{multicol}
\usepackage{wrapfig}
\usepackage{framed}
\usepackage{enumitem}
\usepackage{dsfont, amsmath, amssymb, amsthm}
\usepackage{amsmath}
\usepackage{amssymb}
\usepackage{makecell}
\usepackage[]{algorithm2e}
\usepackage{tikz}
\usepackage{color, colortbl}
\usepackage{verbatim}
\usepackage{threeparttable}

\usepackage{amsmath}
\newcommand{\argmin}{\mathop{\mathrm{argmin}}\limits}

\definecolor{Gray}{gray}{0.9}
\let\OLDthebibliography\thebibliography
\renewcommand\thebibliography[1]{
  \OLDthebibliography{#1}
  \setlength{\parskip}{0pt}
  \setlength{\itemsep}{0pt plus 0.1ex}
}
\usepackage{microtype}

\aclfinalcopy 

\title{Subword Sampling for Low Resource Word Alignment}

\author{Ehsaneddin Asgari\thanks{\mbox{\ \ }  Equal contribution} $^{\dag}$, Masoud Jalili Sabet $^{*,\diamond}$, Philipp Dufter$^\diamond$, Christopher Ringlstetter$^{\dag}$, Hinrich Sch\"{u}tze$^\diamond$\\
$^\diamond$ Center for Information and Language Processing, LMU Munich, Germany \\
$^\dag$ NLP Expert Center, Data:Lab, Volkswagen AG, Munich, Germany\\
{\tt ehsaneddin.asgari@volkswagen.de masoud@cislmu.org}
\\}

\date{}

\begin{document}
\maketitle
\vspace{-0.6cm}
\begin{abstract}
Annotation projection is an important area in NLP that can
greatly contribute to creating language resources for
low-resource languages. Word alignment plays a key role in
this setting. However, most of the existing word alignment
methods are designed for a high resource setting in machine
translation where millions of parallel sentences are
available. This amount reduces to a few thousands of
sentences when dealing with low-resource languages
failing the existing established IBM models.  In this paper, we propose
\emph{subword sampling-based alignment} of text units. This method's hypothesis is that the aggregation of
different granularities of text for certain language pairs
can help word-level alignment. For certain languages for which
gold-standard alignments exist, we propose an iterative
Bayesian optimization framework to optimize selecting
possible subwords from the space of possible subword
representations of the source and target sentences. We show
that the subword sampling method consistently outperforms word-level alignment on six
language pairs: English-German, English-French,
English-Romanian, English-Persian, English-Hindi, and
English-Inuktitut. In addition, we show that the
hyperparameters learned for certain language pairs can be
applied to other languages at no supervision and
consistently improve the alignment results. We observe that using $5K$ parallel sentences together with our proposed subword sampling approach, we obtain similar F1 scores to the use of $100K$'s of parallel sentences in existing word-level fast-align/eflomal alignment methods.\end{abstract}

\section{Introduction}
Annotation projection is an important area in Natural Language Processing (NLP), which aims to exploit existing linguistic resources of a particular language for creating comparable resources in other languages (usually low resource languages) using a mapping of words across languages. More precisely, annotation projection is a specific use of parallel corpora, corpora containing pairs of translated sentences from language $l_s$ to $l_t$.
In annotation projection, a set of labels available for language $l_s$ is projected to language $l_t$ via alignment links (the mapping between words in parallel corpora of $l_s$ and $l_t$).
$L_s$ labels can either be obtained through manual annotation or through an analysis module that may be available for $l_s$, but not for $l_t$.
The label here can be interpreted broadly, including, e.g.,  part of speech
labels, morphological tags and segmentation boundaries,
sense labels, mood labels, event labels, syntactic analysis, and coreference
\cite{yarowsky2001inducing,diab02senseparallel,agic2016multilingual}.

Language resource creation for low-resource languages, for
the purpose of automatic text analysis can create
financial, cultural, scientific, and political value.  For
instance, the creation of a sentiment lexicon for a low
resource language would be an excellent help for customer
reviews analysis in big corporations having branches all
over the world, where 7000 languages are spoken, or such a resource can be used to predict stock market movements from social media in a low resource setting. Furthermore, such resources can contribute to creating knowledge \cite{ostling2015word,asgarischutze2017past} for linguists, which pure scientific value aside, the linguistic
knowledge can be incorporated into machine learning models for natural language understanding
as well.

The mapping between words across languages as a basis for annotation projection is automatically generated using statistical word alignment, modeled on parallel corpora. This means that given parallel corpora for a set of languages and linguistic resources for only one language, we can automatically create resources for the other languages through annotation projection. One of the main challenges for annotation projection is that corpora are often relatively small for low resource languages. The existing IBM-based alignment models
work well for high-resource settings, but they fail in the low-resource case \cite{poerner2018aligning}. The most popular dataset for low resource alignment, the Bible Parallel Corpus, containing a large number (1000+) of languages, are characteristically low-resource, i.e., having only around 5000-10000 parallel sentences per language pair.
This paper aims to introduce a framework to reliably relate linguistic units, words, or subwords, in a low resource parallel corpus, based on sampling from the space of possible subwords.

\section{Methods}
\subsection{Dataset}

We work on the word alignment gold standards from the WPT 2003 and 2005 shared tasks on word alignment. Those language pairs include
French, Hindi, Romanian, and Inuktitut always paired with English.
In addition, we add English-Persian and English-German. As per the standard scenario in the world alignment literature, we compute an alignment model on an independent corpus of training materials. 
To simulate a low-resource scenario, we sample the number of
parallel training sentences down to 5000, except for Hindi with 3000 sentence pairs. This is the order of magnitude in training data
when dealing with low-resource languages contained,  e.g., in the Bible Parallel Corpus. In addition, we experiment on mid-resource cases when using the complete set of available training sentences in English-Romanian, English-Inuktitut, English-Persian, in which their complete set is neither low-resource nor contain more than 1M parallel sentences. See Table \ref{data} for details on the data. 

\begin{table*}
	\centering
	\begin{threeparttable}
		\centering
		\scriptsize
		\begin{tabular}{l||rrrr|rr}
			& Gold &  & && Parallel Training &   \\
			Lang. & Standard & \# Sentences & $|S|$&$|P\setminus S|$&Data & \# Sentences  \\
			\hline
			English-German & EuroParl-based\tnote{a}   &  508 &9612&921&  EuroParl \cite{koehn2005europarl} & 1920k \\
			English-Persian & \cite{tavakoli2014phrase} &  400 &11606&0& TEP \cite{pilevar2011tep} & 600k \\
			English-French & WPT2003, \cite{och2000improved}, & 447 &4038&13400& Hansards \cite{germann2001aligned} & 1130k \\
			English-Hindi & WPT2005\tnote{b} & 90 &1409&0& Emille \cite{mcenery2000emille} & 3k \\
			English-Inuktitut & WPT2005\tnote{b} & 75 & 293 &1679& Legislative Assembly of Nunavut\tnote{b} & 340k \\
			English-Romanian & WPT2005\tnote{b}& 203 &5033&0& Constitution, Newspaper\tnote{b} & 50k \\
		\end{tabular}
		\begin{tablenotes}
			\item[a]  \url{www-i6.informatik.rwth-aachen.de/goldAlignment/} 
			\item[b] \url{http://web.eecs.umich.edu/~mihalcea/wpt05/}
		\end{tablenotes}
	\end{threeparttable}
	\caption{Details on gold standards and training data. $|S|$ is the number of sure edges in the gold standard and $|P\setminus S|$ the number of additional possible edges.  \label{data}}
\end{table*}

\subsection{Evaluation}

We evaluate word alignments with $F_1$ score computed by 
\begin{equation*}
\text{prec} = \frac{|A \cap P|}{|A|},\;
\text{rec} = \frac{|A \cap S|}{|S|},\;
F_1 = \frac{2 \; \text{prec}\; \text{rec}}{\text{prec} + \text{rec}},\\
\end{equation*}
where $|A|$ is the set of predicted alignment edges,  $|S|$ the set of sure and 
 $|P|$ the set of possible alignment edges. Note that $S\subset P$, and both are known from the gold standard.

\subsection{Sentence subword space}
\label{sec:subspace}
For splitting text into subwords, we use Byte-Pair-Encoding by \newcite{sennrich-etal-2016-neural}.
The BPE algorithm for a certain random seed and a given
vocabulary size (analogous to the number of character merging
steps) breaks a sentence into a unique sequence of
subwords. Continuing the merging steps would result in the
enlargement of the subwords, resulting in fewer tokens.

\noindent \textbf{Hypothesis:} Let $S_{pq} =
\bigcup_{j=1}^{N} (s^{(j)}_{p},s^{(j)}_{q})$ be a collection
of $N$ parallel paired sentences in the language pair $l_p$
and $l_q$. We assume that for a certain $S_{pq}$ there exists
an optimal segmentation scheme constructed by accumulation
of different granularities of ($l_p$,$l_q$),
$\xi^{*}$, among all possible segmentation schemes ($\xi$'s),
which depends on the morphological structures of this
language pair.

The space of possible segmentations of a
sentence $s$, denoted as $\Phi_l(s)$ for language $l$,
is created by 
variations in the segmentation by varying the
number of merging
steps.

In this
notation $\Phi_l(s) = \bigcup_{i=1}^{M_l}
\Phi_{l}^{(i)}(s)$, where $\Phi_{l}^{(k)}(s)$ refers to a
specific vocabulary size selection for the segmentation of
$s$ considering the first $k$ merging steps in the BPE
algorithm for language $l$. $M_l$ is the maximum number of merging steps  in $l$. We define $\Phi_{pq}$ as the set
of all possible segmentation pairs in language pair $l_p$
and $l_q$:
\[
\Phi_{pq} = \bigcup_{i=1}^{M_{l1}} \Phi^{(i)}_{p} \times  \bigcup_{i=1}^{M_{l2}} \Phi^{(i)}_{q}
\]



When we deal with a single language, to explore the possible
segmentations, Monte Carlo sampling from $\Phi_{l}$ can be
used to have different views on the segmentation, where the
likelihood of certain segmentation $\Phi_{l}^{(k)}$ is
proportional to the number of sentences affected in the
corpus by introducing the $k^{th}$ merging step, as proposed in
\cite{asgari2019probabilistic,asgari2019life,asgari2019ditaxa} for the segmentation of
protein and DNA sequences. However, in the alignment problem, we deal
with a 2D space (can be represented as a grid as in Figure \ref{fig:bo_eg}) of possibilities for the vocabulary sizes
($\approx$ the number of merging steps in BPE) of $l_p$ and
$l_q$. The inclusion of each cell in this grid introduces new
instances to the parallel corpus, potentially transferring a
low-resource setting to a high-or-mid resource setting. In this
high resource setting, the subwords of a certain sentence
are assigned in $T$ ways (the number of cells we select from
the $\Phi_{pq}$ grid). Finally, to confirm an alignment link at word-level, we set a threshold  $\lambda$; $\lambda$ is the minimum ratio of
subword segmentation is required to confirm a word alignment link. Note
that not necessarily all cells of the grid improve the
alignment, we thus need a strategy to pick a subset of cells
$\xi^{*} \subset \Phi_{pq}$ maximizing the ultimate
alignment score. Having language pairs with ground-truth
alignment, we can solve this problem via hyperparameter
optimization using Bayesian optimization. Subsequently, we
investigate whether applying the same hyperparameters, on
another language pair yields improvements. To solve this
the optimization problem for the supervised case, we propose an
iterative greedy subword sampling algorithm.

\subsection{Iterative subword sampling algorithm}

To maximize the alignment score for the known links (the ground-truth) at the word-level, we are seeking for $\xi^{*}$ a set of cells in the $\Phi_{pq}$ grid, and their corresponding thresholds $\lambda^{*}$ satisfying the following equation:
\[
\xi^{*}, \lambda^{*} = \argmin_{{\footnotesize \xi^{i}, 0 \le\lambda\le1}; i \in \{1,2, .., T\}} -f(\Phi_{pq}, S_{pq}, \mathbf{y_{pq}}),
\]

\noindent where $f$ refers to the alignment F1 score based
on ground-truth, which its underlying alignment model does not have any closed form nor
gradient. $\mathbf{y_{pq}}$ is the ground-truth we have for
the language pair $l_p$ and $l_q$, and $S_{pq}$ refers to the
parallel sentences, which are going to be segmented in $T$
different schemes ($T$ cells from the $\Phi_{pq}$ grid, $0 <
i < T$). These $T$ cells can be selected in any
order. However, to reduce the search space, we propose a
sequential greedy selection of the segmentations $(\xi^{i},
\lambda)$, and solve each step in a Bayesian
optimization framework. The iterative process is detailed in
Algorithm \ref{alg:itsubsam}. The core computation of this
algorithm is   $\xi_{i}, \lambda =\argmin_{\xi_{i},\lambda}
- f(\Phi_{pq}, S_{pq}, \mathbf{y_{pq}}, \xi_{0:i-1})$,
for which the selected vocabulary sizes up to the current iteration ($\xi_{0:i-1}$) are
used for segmentation and the measurement of the alignment score. We
perform Bayesian optimization to find the next optimal
values for  $\xi_{i}$ and $\lambda$. As discussed in
$\S\ref{sec:subspace}$, in the Bayesian optimization, we
explore the cells from the grid of $\Phi_{pq}$ using
logarithmic priors for each of
$\Phi_p$ and $\Phi_q$. We continue this process until the
the moment where introducing more segmentations does not improve
the alignment score, setting an early stopping condition.


\begin{algorithm}
\SetAlgoLined
\KwResult{{\footnotesize graph $G$ of word-aligned sentence pairs}}
 $f1_{prev} = 0$; $i = 0$\; 
 $\xi$ = $<empty>$; {\footnotesize$\%$ history of selected cells}\\
 $\lambda$ = $<empty>$; {\footnotesize$\%$ history of selected $\lambda$'s}\\
 $\delta = +\infty$;  $E$ = early-stopping parameter\;
 \While{$ \exists \delta > 0$ in the last $E$ iterations  }{
  $\xi_{i}, \lambda =\argmin_{\xi_{i},\lambda} - f(\Phi_{pq}, S, y,\xi, \lambda)$\;
  $\xi.push(\xi_{i})$\;
  $\lambda.push(\lambda)$\;
  $f1^{*} = f(S, y, \xi, \lambda)$\;
  $\delta = f1^{*} - f1_{prev}$\;
  $f1_{prev} = f1^{*}$\;
  $i = i + 1$\;
 }
 G = $alignment(segment(S, \xi, \lambda ))$
 \caption{Iterative subword sampling\label{alg:itsubsam}}
\end{algorithm}

\subsection{Intuition behind the use of logarithmic priors for the vocabulary size}

\begin{figure}[ht!]
 {\centering
\includegraphics[trim={0cm 0cm 0cm 0cm},clip,width=1\columnwidth]{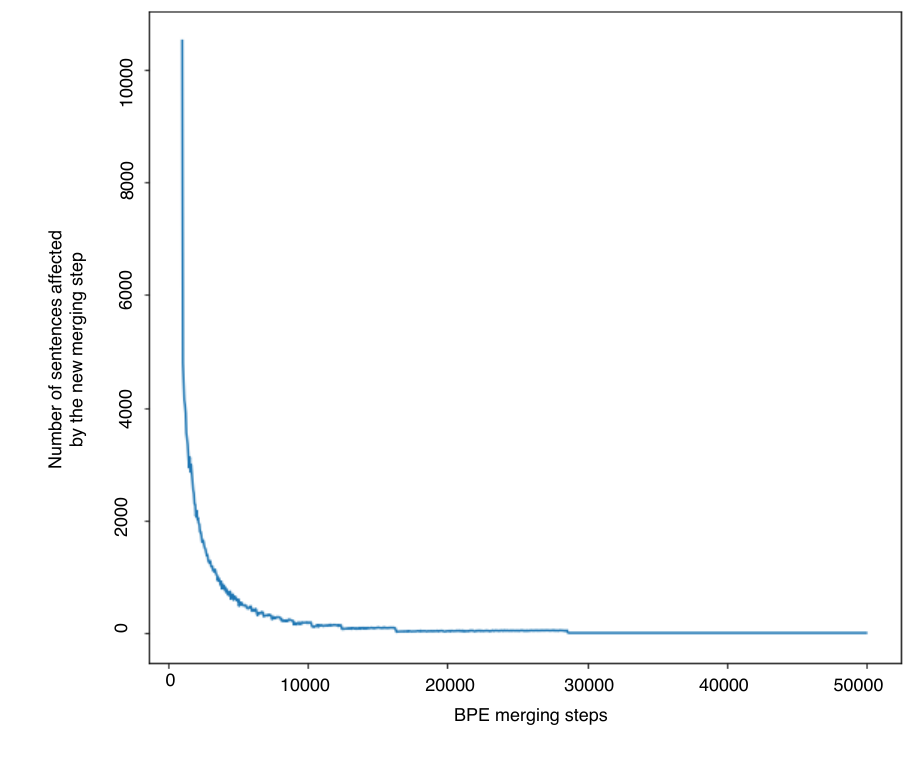} \\} \captionof{figure}{An example of English BPE on a collection of 10000 sentences. This diagram shows that with introducing new merging steps how many sentences are going to be affected .\label{fig:bpe_size}}
\end{figure}

Figure \ref{fig:bpe_size} provides an intuition behind the use of logarithmic priors. This diagram shows that by introducing a new merging step (increasing the vocabulary size by one) in the BPE algorithm, which portion of sequences are affected. As proposed for protein sequences \cite{asgari2019probabilistic}, this can be served as an approximation for the relative likelihood of including a merging step (which is analogous to introducing a new subword).

\subsection{Subword sampling in other languages}
After training how to choose $\xi^{*}, \lambda^{*}$ for a particular language pair, we apply the same vocabulary settings on new language pairs and evaluate the resulting alignment scores. This would help us in the investigation of the generalizability of a language pair on other language pairs.

\subsection{Experimental Setup}

\subsubsection*{Evaluate Low-Resource Alignment}
Since the main motivation of subword sampling alignment is for the low-resource scenario\footnote{The language itself is not necessarily a low-resource language, but the number of sentence pairs is relatively low (less than 10K)},  we first evaluate the method for an analogous use case, where only a few thousands of parallel sentences are available, similar to the Bible Parallel Corpus of 1000+ languages.
To produce a similar scenario for the evaluation, we get samples of 5K aligned sentences from the training datasets of English-German, English-French, English-Romanian, English-Persian, English-Hindi, and English-Inuktitut and concatenate the gold-standard datasets to them.
The statistical word aligners generate forward, and backward alignments need a post-processing step of symmetrization \cite{koehn2010statistical}.
We compared intersection and grow-diag-final-and (GDFA), which produce comparable results in terms of F1 score, and the intersection method having a higher precision.
Since the final alignments are produced from the aggregation of all segmentations' alignments, the intersection method with higher precision is a proper candidate. Thus, we use the intersection method throughout the experiments.

For each language pair, we evaluate the word-level alignment, as well as the Bayesian optimization subword sampling.
In addition, in order to investigate how the vocabulary size of a particular language pair generalizes to the other language pairs, we also evaluate the optimized  $\xi^{*}_{l_1,l_2}, \lambda^{*}_{l_1,l_2}$ for each pair on all other language pairs.

\subsubsection*{Evaluate Mid-Resource Alignment}

In addition to the low-resource alignment, we evaluate our approach against the word-level alignment of fast-align and eflomal in the mid-resource scenario (having less than 1M sentence pairs). Therefore, From the six language pairs with gold-standard alignment, we select English-Persian, English-Inuktitut, and English Romanian, containing 600k, 340k, 50k sentence pairs, respectively. For each language pair, we use the vocabulary sizes optimized in the low-resource alignment experiment.

\section{Results}

\subsection{Iterative Subword Sampling}
\begin{figure*}[ht!]
  {\centering
 \includegraphics[trim={1cm 1cm 1cm 0cm},clip,width=1.5\columnwidth]{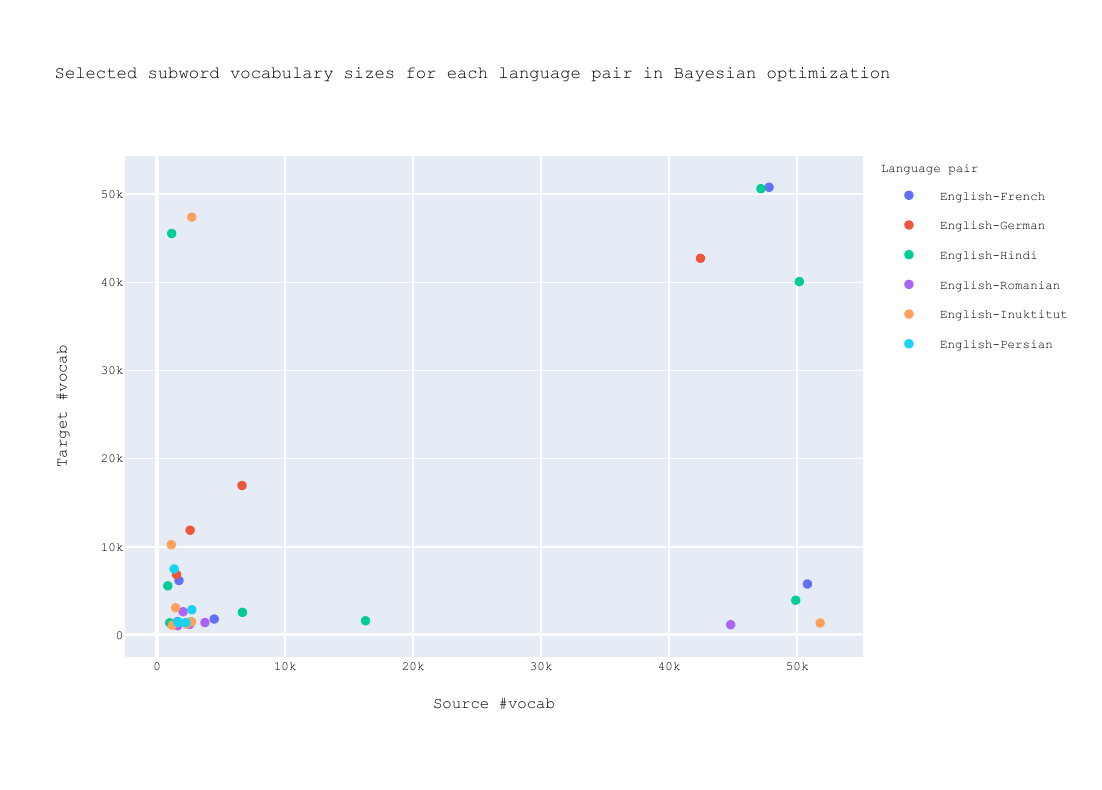}\\}
 \captionof{figure}{Selected subword vocabulary sizes within Bayesian optimization for each language pairs of English-German, English-French, English-Romanian, English-Persian, English-Hindi, and English-Inuktitut.\label{fig:selected}}
\end{figure*}

\begin{figure*}[ht!]
\centering 
\begin{minipage}{.53\textwidth}
  {\centering
 \includegraphics[trim={0cm 0cm 0cm 0cm},clip,width=0.85\columnwidth]{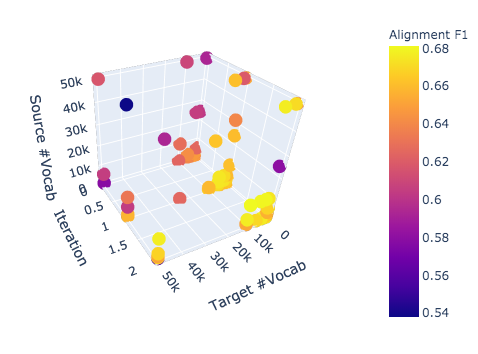}\\}
 \captionof{figure}{The space of $\Phi_{pq}$ that is explored in the Bayesian optimization in the first 3 iterations. The exploring steps are colored with their alignment F1 scores.\label{fig:bo}}
  \label{fig:test1}
\end{minipage}%
\hspace{0.3cm}
\begin{minipage}{.38\textwidth}
  {\centering
\includegraphics[trim={0cm 0cm 0cm 0cm},clip,width=0.85\columnwidth]{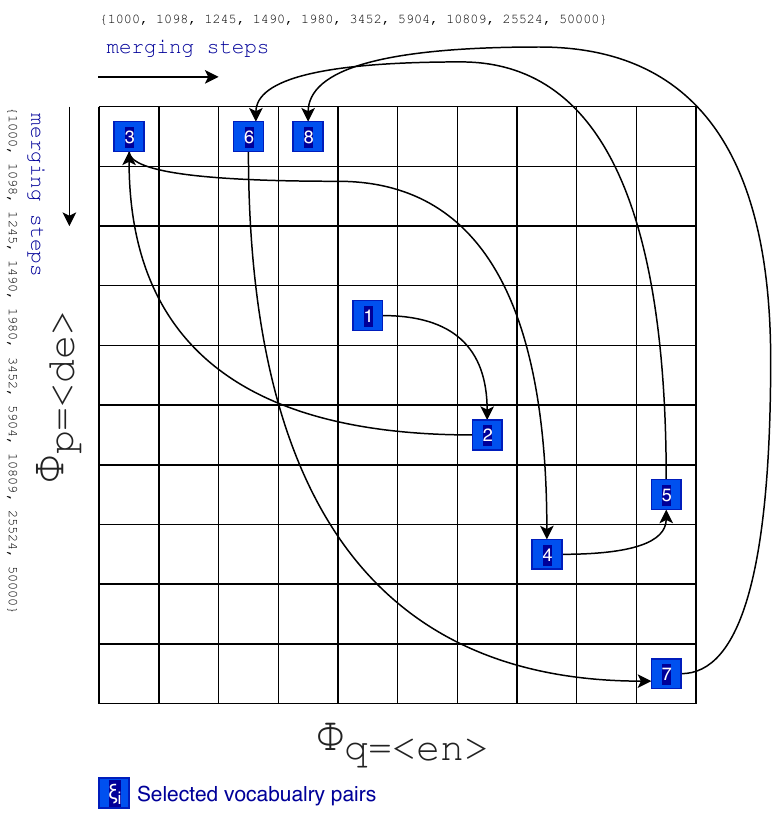} \\} \captionof{figure}{An example of the space $\Phi_{pq}$ for English and German and the selected cells by the Bayesian optimization.\label{fig:bo_eg}}
  \label{fig:test2}
\end{minipage}
\end{figure*}

An example space of $\Phi_{pq}$ (for English-German) that is
explored in the Bayesian optimization to find the $\xi^{*}$
is shown in Figure \ref{fig:bo}, a 2D representation of the
selected cells and the order of selection by Bayesian
optimization on the English-German corpus is provided in Figure
\ref{fig:bo_eg}. We observe that the new segmentation in each iteration consistently improves the alignment scores in the next iteration. Furthermore, as may be expected, the sampled vocabulary sizes are mainly chosen from the lower sizes, i.e., affecting more sentences (Figure \ref{fig:bpe_size}). All studied language pairs show similar behaviour in selecting subword vocabulary sizes (Figure \ref{fig:selected}).

\subsection{Low-Resource Alignment Results}
 
 Table \ref{tab:prediction_res} shows $F_1$ scores of alignment across six language pairs in the low-resource alignment (having a maximum set of 5K aligned sentence pairs). This table compares the word-level and subword-level alignments as well as the generalizability of the $\xi^{*}_{l_1,l_2}, \lambda^{*}_{l_1,l_2}$ on the other language pairs. 
Interestingly, across all language pairs, we observe improvements of 1.4 to 8.0 percentage points in the alignment F1 score in comparison with the word-level alignment. Subword-sampling optimized on a specific pair  consistently improves the word-level alignment also of the other languages. Certain language pairs, like Romanian-English and Hindi-English, proved better generalizability when applied to the other language pairs. This result suggests that although the gold standard is
decisive for a significant improvement of the alignment through optimizing the vocabulary sizes, then optimal vocabulary sizes trained on different language pairs
(potentially with similar morphological complexity) can be efficiently applied to increase the alignment performance for a new language pair .

\begin{table*}[ht!]
{\centering
\resizebox{2\columnwidth}{!}{

\begin{tabular}{l|c|c|c|c|c|c|}
\cline{2-7}
                                                                       & English-German           & English-French           & English-Romanian                                & English-Persian          & English-Hindi            & English-Inuktitut        \\ \hline\hline
\multicolumn{1}{|c|}{word-level}                                       & 0.685                    & 0.897                    & 0.616                                           & 0.527                    & 0.508                    & 0.771                    \\ \hline
\multicolumn{1}{|l|}{Vocab-size Sampling Optimization}                 & 0.746                    & 0.913                    & 0.662                                           & 0.579                    & 0.548                    & 0.858*                 \\ \hline\hline
\multicolumn{1}{|l|}{Apply \textless{}English-German\textgreater{}}    & \cellcolor[HTML]{C0C0C0} & 0.899                    & 0.645                                           & 0.560                     & 0.532                    & 0.845*                    \\ \hline
\multicolumn{1}{|l|}{Apply \textless{}English-French\textgreater{}}    & 0.743                    & \cellcolor[HTML]{C0C0C0} & 0.651                                           & 0.541                    & 0.524                    & \textbf{0.853}                    \\ \hline
\multicolumn{1}{|l|}{Apply \textless{}English-Romanian\textgreater{}}  & \textbf{0.745}           & 0.905                    & \cellcolor[HTML]{C0C0C0}{\color[HTML]{FFFFFF} } & 0.579                    & \textbf{0.547}                    & 0.821             \\ \hline
\multicolumn{1}{|l|}{Apply \textless{}English-Persian\textgreater{}}   & 0.743                    & 0.908           & \textbf{0.663}                                           & \cellcolor[HTML]{C0C0C0} & 0.521                    & 0.816                 \\ \hline
\multicolumn{1}{|l|}{Apply \textless{}English-Hindi\textgreater{}}     & 0.742                    & 0.904                    & \textbf{0.663}                                           & \textbf{0.580}            & \cellcolor[HTML]{C0C0C0} & 0.804                    \\ \hline
\multicolumn{1}{|l|}{Apply \textless{}English-Inuktitut\textgreater{}} & 0.744                    & \textbf{0.914}                    & 0.655                                  & 0.530            & 0.529           & \cellcolor[HTML]{C0C0C0} \\ \hline
\end{tabular}
}}
\caption{The alignment performances (in terms of F1 score) of six language pairs in the low-resource scenario, where the subword sampling and word-level alignments are compared. In addition, the results on applying the hyper-parameters of language pairs on all other pairs are also provided. We experimented systematically on the use of both eflomal and fast-align for every setting. However, for simplicity, in each cell, the best performance of fast-align and eflomal is reported. With the exception of the marked F1
's with \textbf{*}, the best results obtained using eflomal method for all the alignments.}
 \label{tab:prediction_res}
\end{table*}

\subsection{Mid-Resource Alignment Results}
$F_1$ scores of alignment across three language pairs in the mid-resource alignment (having less than 1M aligned sentence pairs) is shown in 
Table \ref{tab:prediction_medsize}.  This table compares the word-level and subword-level alignments and the generalizability of the hyper-parameter optimized on other language pairs in low-resource for a mid-resource setting. Again, across all language pairs, we observe improvements of 2.7 to 7  percentage points in the alignment F1 score compared to the word-level alignment. Interestingly, the F1 we achieved, using 5K parallel sentences and subword sampling, is similar to the word-level F1 score of English-Inuktitut using 240K parallel sentences and the word-level F1 score of English-Persian using 600K parallel sentences.

\begin{table*}[ht!]
{\centering
\resizebox{2\columnwidth}{!}{
\begin{tabular}{
>{\columncolor[HTML]{FFFFFF}}l |
>{\columncolor[HTML]{FFFFFF}}c ||
>{\columncolor[HTML]{FFFFFF}}c |
>{\columncolor[HTML]{FFFFFF}}c |
>{\columncolor[HTML]{FFFFFF}}c |}
\cline{2-5}
                                                                                                                 & Alignment method & English-Romanian (50K)       & English-Inuktitut (340K) & English-Persian (600K) \\ \hline
\multicolumn{1}{|l|}{\cellcolor[HTML]{FFFFFF}}                                                                   & fast-align       & 0.643                        & 0.794                    & 0.552                  \\ \cline{2-5} 
\multicolumn{1}{|l|}{\multirow{-2}{*}{\cellcolor[HTML]{FFFFFF}word-level}}                                       & eflomal          & 0.692                        & 0.864                    & 0.58                   \\ \hline\hline
\multicolumn{1}{|l|}{\cellcolor[HTML]{FFFFFF}}                                                                   & fast-align       & 0.667                        & \textbf{0.915}           & 0.525                  \\ \cline{2-5} 
\multicolumn{1}{|l|}{\multirow{-2}{*}{\cellcolor[HTML]{FFFFFF}Apply \textless{}English-German\textgreater{} parameters}}    & eflomal          & 0.715                        & 0.849                    & 0.638                  \\ \hline\hline
\multicolumn{1}{|l|}{\cellcolor[HTML]{FFFFFF}}                                                                   & fast-align       & 0.664                        & 0.913                    & 0.534                  \\ \cline{2-5} 
\multicolumn{1}{|l|}{\multirow{-2}{*}{\cellcolor[HTML]{FFFFFF}Apply \textless{}English-French\textgreater{} parameters}}    & eflomal          & 0.709                        & 0.885                    & 0.647                  \\ \hline\hline
\multicolumn{1}{|l|}{\cellcolor[HTML]{FFFFFF}}                                                                   & fast-align       & {\color[HTML]{000000} 0.663} & 0.873           & 0.534                  \\ \cline{2-5} 
\multicolumn{1}{|l|}{\multirow{-2}{*}{\cellcolor[HTML]{FFFFFF}Apply \textless{}English-Romanian\textgreater{} parameters}}  & eflomal          & 0.712                        & 0.826                    & 0.587                  \\ \hline\hline
\multicolumn{1}{|l|}{\cellcolor[HTML]{FFFFFF}}                                                                   & fast-align       & 0.677                        & 0.897                    & 0.562                  \\ \cline{2-5} 
\multicolumn{1}{|l|}{\multirow{-2}{*}{\cellcolor[HTML]{FFFFFF}Apply \textless{}English-Persian\textgreater{} parameters}}   & eflomal          & 0.711                        & 0.812                    & 0.587                  \\ \hline\hline
\multicolumn{1}{|l|}{\cellcolor[HTML]{FFFFFF}}                                                                   & fast-align       & 0.659                        & 0.865                    & 0.521                  \\ \cline{2-5} 
\multicolumn{1}{|l|}{\multirow{-2}{*}{\cellcolor[HTML]{FFFFFF}Apply \textless{}English-Hindi\textgreater{} parameters}}     & eflomal          & \textbf{0.719}               & 0.813                    & 0.606                  \\ \hline\hline
\multicolumn{1}{|l|}{\cellcolor[HTML]{FFFFFF}}                                                                   & fast-align       & 0.654               & 0.911                    & 0.519                  \\ \cline{2-5} 
\multicolumn{1}{|l|}{\multirow{-2}{*}{\cellcolor[HTML]{FFFFFF}Apply \textless{}English-Inuktitut\textgreater{} parameters}} & eflomal          & 0.71                         & 0.898                    & \textbf{0.65}          \\ \hline
\end{tabular}}}
\caption{The alignment performances (in terms of F1 score) of three language pairs in mid-resource scenario, where the subword sampling and word-level alignments are compared. In addition, the results on applying the hyper-parameters of language pairs on all other pairs are also provided. For each setting, both eflomal and fast-align results are reported.}
 \label{tab:prediction_medsize}
\end{table*}


\subsection{Qualitative Analysis}

\begin{figure*}[t]
    \centering
    \includegraphics[width=0.32\linewidth]{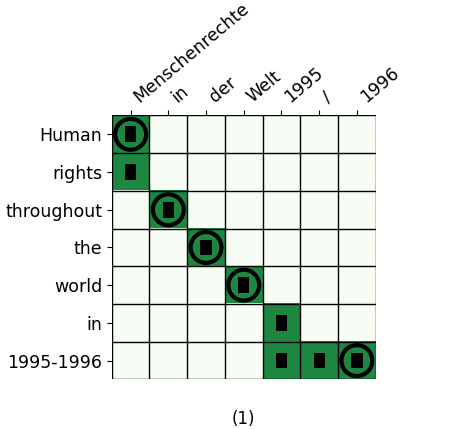}
    \includegraphics[width=0.32\linewidth]{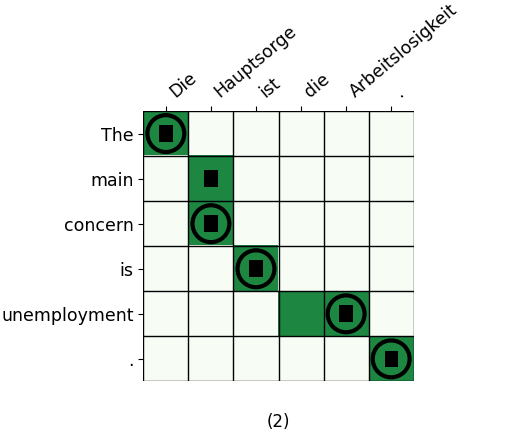}
    \includegraphics[width=0.32\linewidth]{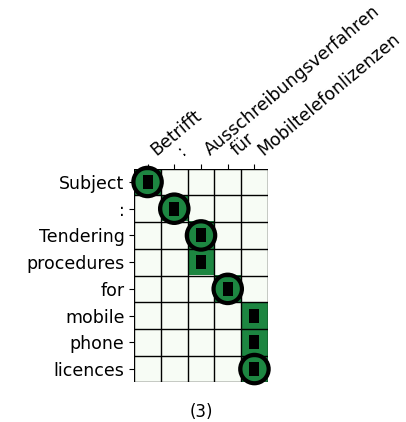}\\
    \vspace{0.3cm}
    \includegraphics[width=0.25\linewidth]{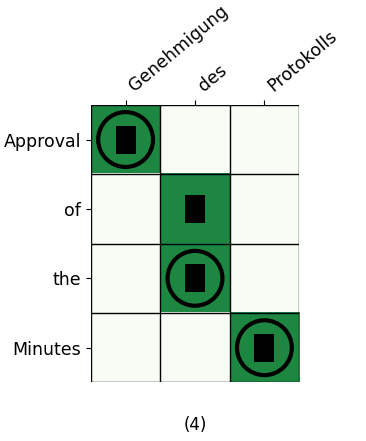}
    \includegraphics[width=0.28\linewidth]{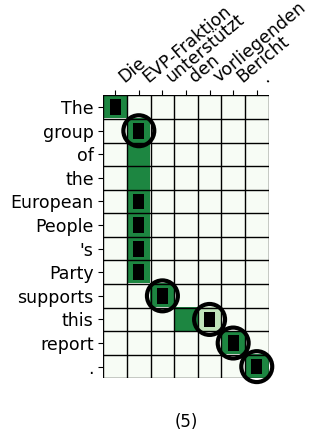}
    \includegraphics[width=0.32\linewidth]{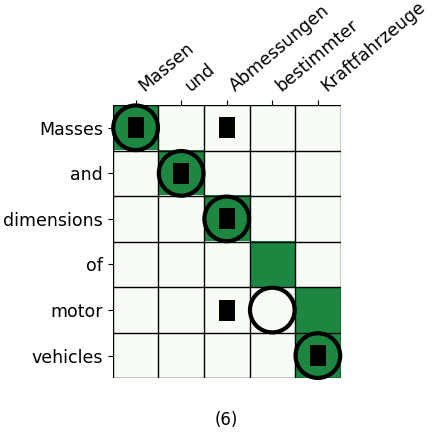}
    \caption{Examples from the English-German gold standard. Dark green represent sure alignment edges. Light green possible edges (only on edge in Example 5). Circles are edges predicted by word-level eflomal, boxes are predicted when applying our proposed subword sampling with eflomal.}
    \label{fig:examples}
\end{figure*}

We performed a qualitative analysis for English-German and showed six examples in Figure \ref{fig:examples}. We observed two sources of improvements: i) Compounds, which are frequent in German, obtain better alignments and ii) As we aggregate alignment edges through a $\lambda$-weight vote, we observe an ``ensembling'' effect which mainly affects fertility. Examples 1, 2 and 3 show the compound effect: ``Menschenrechte'' is correctly aligned to ``Human rights'' only when using sampling optimization. Similarly, ``Mobiltelefonlizenzen'' gets successfully aligned to ``mobile phone license'' whereas pure eflomal only aligns it to ``licenses''. In addition, the differently formatted year numbers in Example 2 are easy to align once subword sampling is used. Examples 4 and 5 show the presumed ensembling effect. We hypothesize that 
``EVP'' is aligned to different words in the English sentence across different subword samples. Once aggregated, ``EVP-Fraktion'' has high fertility, which is useful in this scenario. Similarly, ``des'' (meaning ``of the'') receives a better alignment through subword sampling as some models align ``des'' to ``of'' and some others to ``the''.
Subword sampling cannot resolve all errors of eflomal and can also be harmful in rare cases. Example 6 shows a case where the word ``Abmessungen'' (``dimensions'' or ``measurements'') obtains two incorrect alignment edges, presumably because it frequently gets split into subwords like ``Ab'' or ``ungen'' which carry only little semantic information.

\section{Related Work}
\emph{Classical models.} Statistical word alignment methods
(e.g., GIZA++~\cite{och2000improved},
fast-align~\cite{dyer2013simple}, eflomal~\cite{ostling2016efficient}) are mostly based on
\emph{IBM models}~\cite{brown1993mathematics}, which are
generative models describing how a source language sentence
$S$ generates a target language sentence $T$ using
alignment latent variables. These models use an expectation
maximization (EM) algorithm to train the alignment and only
require sentence-aligned parallel
corpora.

\emph{Neural models.} In 2014, seq2seq
recurrent neural network (RNN) models introduced for machine translation
providing an end-to-end translation
framework~\cite{sutskever2014sequence}. Attention 
was a key component to improve such
models~\cite{bahdanau2014neural,luong2015effective}. 
Two modifications to attention were proposed to improve
the quality of underlying alignment and consequently, the quality of translation.
(i) Model guided
alignment training is
introduced~\cite{chen2016guided,mi2016supervised,garg2019jointly,stengel2019discriminative} where the
cross-entropy between attention weights and the alignment
coming from an IBM model (GIZA++) or a manual gold standard is used as an additional
cost function. \newcite{garg2019jointly} find that operating at the
subword-level can be beneficial for alignment
models. Note that they only consider a single subword segmentation. (ii)
A disadvantage of neural architectures in comparison with IBM
models in producing alignments is that in the neural model
the attention weights have only observed the previous target
words; in contrast, the IBM models benefit from full observation of
the target sentence in alignment generation.
\emph{Target foresight} \cite{peter2017generating} improves translation by considering the target word of the current decoding step
as an additional input to the attention
calculation. The main purpose of the above-mentioned alignment structures has been to improve translation quality. In contrast, our main motivation is providing a framework to reliably
relate linguistic units, words, or subwords in parallel
corpora, which can be used in linguistic resource creation
\cite{agic2016multilingual,asgari2020unisent} and
typological analysis
\cite{ostling2015word,asgarischutze2017past}. The above
mentioned methods work well for the large parallel corpora, but they fail when parallel sentences are scarce. Insufficiency of parallel sentences is usually the case for low-resource languages, which are usually the most interesting scenarios for linguistic resource creation and linguistic analysis \cite{cieri2016selection}. 

\emph{Low-resource alignment models.} The most
popular dataset for low resource alignment is the Bible Parallel Corpus containing a large number (1000+) of
languages, but are characteristically low-resource, i.e.,
have little text per language \cite{mayer2014creating}. Some
recent work touched upon this problem using unsupervised
cross-lingual embeddings and a monogamy objective
\cite{poerner2018aligning}. However, this method could not improve the fast-align results for the parallel corpora containing more than 250 sentences. We showed that our method improves the fast-align and eflomal on six language pairs consistently on the size of 5000K parallel sentences, in the range of parallel sentences of 1000+ languages in BPC, the most interesting parallel corpora for the low-resource scenario (in terms of the number of covered languages). Our proposed method improved the mid-resource alignments (50K-600K parallel sentences) as well.


\emph{Subword sampling.} The use of multiple
subword candidates has improved the machine translation
performance \cite{kudo2018subword}.  BPE-Dropout
\cite{provilkov2020bpe} followed the same idea, introducing
dropout in the merging steps of a fixed BPE to create
multiple segmentations. The probabilistic use of multiple subword candidates has been proposed to segmentation protein sequences \cite{asgari2019probabilistic}.
We use the inspiration from the latter approach for the word-alignment of parallel sequences of language pairs, using a multitude of possible subword segmentations.


\section{Conclusion}
Motivated by the important NLP area of annotation projection, used to create linguistic resources/knowledge in the low-resource languages, we proposed subword sampling-based alignment of text units. This method's hypothesis is that the aggregation of different granularities of text for specific language pairs can help with word-level alignment. For individual languages where a gold-standard alignment corpus exists, we proposed an iterative Bayesian optimization framework to optimize selecting subwords from the space of possible BPE representations of the source and target sentences. We showed that the
subword sampling method consistently outperforms the pure word-level alignment on six language pairs of English-German, English-French, English-Romanian, English-Persian, English-Hindi,
and English-Inuktitut in a low-resource scenario. Although the subword samples are selected in a supervised manner, we show that the hyperparameters can fruitfully be used for other language pairs with no supervision and consistently
improve the alignment results. We showed that using 5K parallel sentences together with our proposed subword sampling approach, we obtain similar F1 scores to the use of 340K and 600K parallel sentences and word-level alignment in English-Inuktitut and English-Persian, respectively. The proposed method can efficiently improve the creation of linguistic resources (POS tagging, sentiment lexicon, etc.) for low-resource languages, where only a few thousand parallel sentences are available.

\section*{Acknowledgment}

This work has been funded by the German Federal Ministry of Education and
Research (BMBF) under Grant No. 01IS18036A. The authors of this work take full responsibility for its content.


\end{document}